\pdfoutput=1

\documentclass{article}

\usepackage{microtype}
\usepackage{graphicx}
\usepackage{subcaption}
\usepackage{booktabs} 
\usepackage{multirow}

\usepackage[hypertexnames=false]{hyperref}
\hypersetup{pdfborder={0 0 0}}

\usepackage[accepted]{icml2026}
\makeatletter
\renewcommand{\ICML@appearing}{}
\makeatother

\usepackage{amsmath}
\usepackage{amssymb}
\usepackage{mathtools}
\usepackage{amsthm}
\usepackage{enumitem}

\usepackage[capitalize,noabbrev]{cleveref}

\graphicspath{{images/}}

\icmltitlerunning{CryptoBench: A Dynamic Benchmark for Expert-Level Evaluation of LLM Agents in Cryptocurrency: A Dynamic Benchmark for Expert-Level Evaluation of LLM Agents in Cryptocurrency}

\makeatletter
\AtBeginDocument{%
  \gdef\@icmltitlerunning{CryptoBench: A Dynamic Benchmark for Expert-Level Evaluation of LLM Agents in Cryptocurrency}%
  \fancyhead[C]{\small\bf CryptoBench: A Dynamic Benchmark for Expert-Level Evaluation of LLM Agents in Cryptocurrency}%
}
\makeatother

\begin{document}

\twocolumn[
  \icmltitle{CryptoBench: A Dynamic Benchmark for Expert-Level Evaluation of LLM Agents in Cryptocurrency}


  \icmlsetsymbol{equal}{*}

  \begin{icmlauthorlist}
    \icmlauthor{Jiacheng Guo}{equal,princeton}
    \icmlauthor{Suozhi Huang}{equal,princeton}
    \icmlauthor{Zixin Yao}{equal,princeton}
    \icmlauthor{Yifan Zhang}{equal,princeton}
    \icmlauthor{Yifu Lu}{equal,princeton}
    \icmlauthor{Jiashuo Liu}{equal,other}
    \icmlauthor{Zihao Li}{princeton}
    \icmlauthor{Nicholas Deng}{other}
    \icmlauthor{Qixin Xiao}{umich}
    \icmlauthor{Jia Tian}{pyra}
    \icmlauthor{Kanghong Zhan}{berkeley}
    \icmlauthor{Tianyi Li}{deepreach}
    \icmlauthor{Xiaochen Liu}{deepreach}
    \icmlauthor{Jason Ge}{zenith}
    \icmlauthor{Chaoyang He}{chainopera}
    \icmlauthor{Kaixuan Huang}{princeton}
    \icmlauthor{Lin F. Yang}{ucla}
    \icmlauthor{Wenhao Huang}{other}
    \icmlauthor{Mengdi Wang}{princeton}
  \end{icmlauthorlist}

  \icmlaffiliation{princeton}{Princeton University}
  \icmlaffiliation{pyra}{Pyra.io}
  \icmlaffiliation{deepreach}{DeepReach.ai}
  \icmlaffiliation{zenith}{Zenith Lab}
  \icmlaffiliation{chainopera}{ChainOpera AI}
  \icmlaffiliation{ucla}{University of California, Los Angeles}
  \icmlaffiliation{berkeley}{University of California, Berkeley}
  \icmlaffiliation{umich}{University of Michigan}
  \icmlaffiliation{other}{\phantom{.}}

  \icmlcorrespondingauthor{Jiacheng Guo}{jiacheng.guo@princeton.edu}

  \icmlkeywords{Machine Learning, LLM Agents, Cryptocurrency, Benchmark}

  \vskip 0.3in
]

\printAffiliationsAndNotice{\icmlEqualContribution}

\begin{abstract}
We introduce CryptoBench, the first expert-curated, dynamic benchmark designed to evaluate LLM agents on real-world cryptocurrency analysis tasks. Unlike general-purpose agent benchmarks~\cite{wei2025browsecomp, mialon2023gaia}, professional crypto analysis presents unique challenges: \emph{extreme time-sensitivity}, \emph{adversarial information environments}, and the need to synthesize data from \emph{diverse specialized sources} such as on-chain intelligence platforms and real-time DeFi dashboards.
To address these challenges, we construct a live benchmark featuring 50 monthly questions designed by crypto-native professionals to mirror actual analyst workflows. Tasks are organized into a four-quadrant system (Simple/Complex $\times$ Retrieval/Prediction), enabling precise assessment of both data-gathering and analytical capabilities.
Our evaluation of ten state-of-the-art LLMs, both directly and within an agentic framework, reveals a critical \textit{retrieval-prediction imbalance}: leading models demonstrate strong performance on data retrieval but exhibit pronounced weakness in predictive analysis. This highlights a problematic tendency for agents to appear factually grounded while lacking the deeper analytical capabilities required for expert-level performance.
\end{abstract}

\section{Introduction}
\label{sec:intro}

\begin{figure*}[t]
  \centering
  \includegraphics[width=0.9\textwidth]{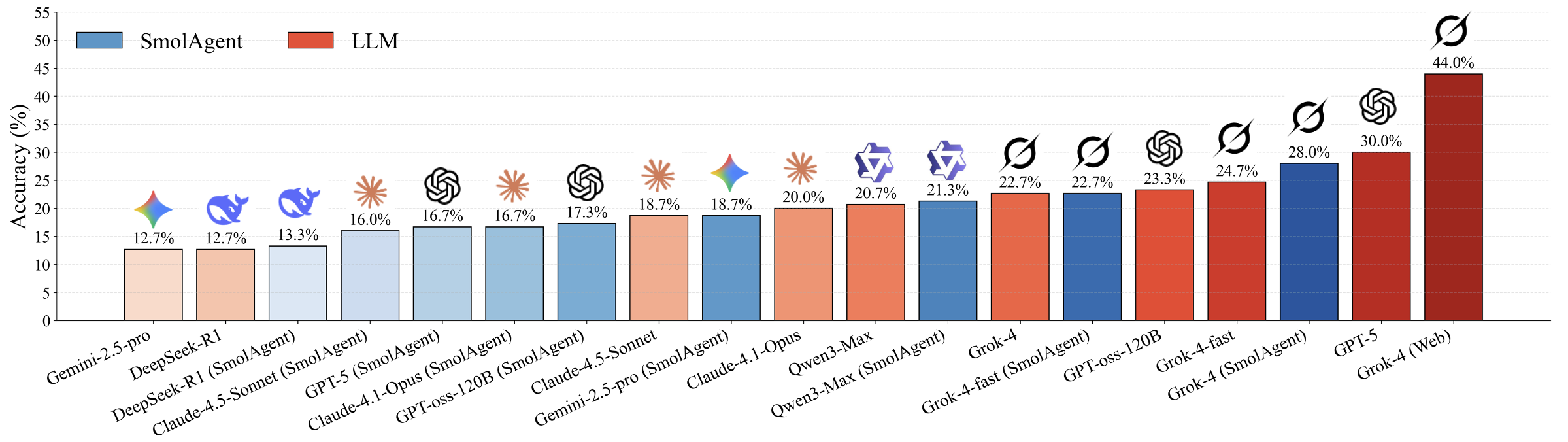}
  \caption{Combined Evaluation Scores of LLM and SmolAgent Frameworks between October $12^{th}$ to November $11^{th}$. Red execution through direct LLM evaluation, while blue bars represent performance within the SmolAgent framework.}
  \label{fig:combined_eval_scores}
\end{figure*}

Large Language Model (LLM) agents mark a transition from passive instruction-following systing systems to autonomous entities capable of multi-step reasoning, tool use, and action reasoning, tool use, and action execution \cite{yao2022react, park2023generative}. This evolution is particularly consequential inis evolution is particularly consequential in finance, where the ability to rapidly synthesize heterogeneous information and act under uncertainty directly determines economic outcomes.

Among financial domains, cryptocurrency markets constitute an especially demanding testbed for agentic capabilities. Crypto markets operate continuously and are characterized by real-time data streams, adversarial behavior, and highly unstructured information sources, including on-chain transaction graphs, protocol dashboards, and social media discourse.

Despite this need, existing agent benchmarks are poorly suited for evaluating expert-level cryptocurrency analysis. General-purpose benchmarks such as WebArena \cite{zhou2024webarena}, GAIA \cite{mialon2023gaia}, and AgentBench \cite{liu2023agentbench} primarily emphasize static environments, historical data access, or factual extraction, and lack the domain-specific data sources and adversarial dynamics inherent to crypto markets.

To address this gap, we introduce \textbf{CryptoBench}, the first benchmark designed to rigorously evaluate LLM agents on realistic, expert-level cryptocurrency tasks. CryptoBench is built around three core design principles: \textit{(1) expert-curated tasks} reflecting professional analyst workflows, \textit{(2) a dynamic structure} that supports continuous updates in rapidly evolving markets, and \textit{(3) real-world execution} through direct interaction with live, specialized crypto platforms.

Tasks in CryptoBench are organized into four categories: Simple Retrieval, Complex Retrieval, Simple Prediction, and Complex Prediction, enabling fine-grained assessment of agent capabilities. Using this framework, we uncover a pronounced \textit{retrieval---prediction imbalance}dels perform competitively on retrieval tasks, but exhibit near-complete failure in predictive reasoning. Furthermore, we find that incorporating an agentic framework substantially alters model performance rankings, indicating that raw model capability does not directly translate into effective agentic execution.

These findings are summarized in Figure~\ref{fig:combined_eval_scores}. Table~\ref{tab:crypto_benchmark_comparison} situates CryptoBench within the broader landscape of existing benchmarks. Additional background and detailed experimental results are provided in Appendix~\ref{sec:appendix}.


\begin{table}[t]
\centering
\caption{Comparison with Previous Benchmarks for Cryptocurrency analysis. A \checkmark in the On-chain analysis column indicates support for on-chain data, though it may not be updated regularly.}
\label{tab:crypto_benchmark_comparison}
\resizebox{\columnwidth}{!}{%
\begin{tabular}{@{}lccccc@{}}
\toprule
\textbf{Benchmark} & \textbf{On-chain} & \textbf{Time} & \textbf{Env.} & \textbf{Freq.} & \textbf{Auto} \\ \midrule
 CAIA \cite{dai2026hallucinationcostsmillionsbenchmarking} & \checkmark & Past & Real & Once & $\times$ \\
AMA \cite{qian2025agentstradelivemultimarket} & $\times$ & Cur./Fut. & Real & Daily & \checkmark \\
InvestorBench \cite{li2025investorbench} & $\times$ & Past & Sim. & Once & $\times$ \\
FutureX \cite{zeng2025futurex} & $\times$ & Future & Real & Weekly & \checkmark \\
CryptoTrade \cite{li2024cryptotrade} & \checkmark & Past & Sim. & Once & $\times$ \\ \midrule
\textbf{Ours} & \checkmark & All & Real & Monthly & \checkmark \\ \bottomrule
\end{tabular}
}
\end{table}

\section{Related Work}
\label{sec:related_work}

\textbf{Agentic Benchmarks.}
A growing body of work has proposed benchmarks for evaluating the general capabilities of LLM agents. Representative examples include WebArena \cite{zhou2024webarena} and BrowseComp \cite{wei2025browsecomp}, which focus on web navigation and information retrieval, as well as GAIA \cite{mialon2023gaia} and AgentBench \cite{liu2023agentbench}, which evaluate agents across diverse environments and task types. SWE-bench \cite{jimenez2023swe} assesses agent performance on real-world software engineering tasks. While these benchmarks provide valuable insights into general agentic behavior, they are intentionally domain-agnostic and do not capture the specialized data sources, domain knowledge, and analytical workflows required in financial settings \cite{bigeard2025finance, li2025investorbench}.

\textbf{Time-Sensitive Benchmarks.}
Recent work has highlighted the importance of evaluating models under evolving, time-sensitive conditions. Benchmarks such as FreshQA \cite{vu2024freshllms}, LiveBench \cite{white2024livebench}, and TimE \cite{wei2025time} incorporate temporal dynamics to assess up-to-date knowledge and prediction over time. CryptoBench extends this line of research by combining real-time information access with a benchmark design that is itself continuously updated, ensuring resistance to data contamination.

\textbf{Future Prediction Benchmarks.}
Evaluating the forecasting and future prediction capabilities of LLMs and agents has become an active research area. Prior work has explored real-world forecasting accuracy \cite{lu2025evaluating}, prediction markets, and dynamic future-oriented benchmarks such as FutureX \cite{zeng2025futurex} and ForecastBench \cite{karger2024forecastbench}. Other benchmarks target future event prediction across broader domains \cite{guan2024openep, nako2025navigating}. CryptoBench builds upon these efforts by focusing specifically on prediction tasks grounded in the cryptocurrency ecosystem, where volatility, adversarial behavior, and domain-specific metrics pose unique challenges \cite{lu2025bizfinbenchbusinessdrivenrealworldfinancial, mateega2025financeqa}. Extended discussion of related work is provided in Appendix~\ref{sec:extended_related_work}.

\section{The CryptoBench Benchmark}
\label{sec:benchmark_design}

\begin{figure*}[t]
    \centering
    \begin{subfigure}{0.48\textwidth}
        \includegraphics[width=\linewidth]{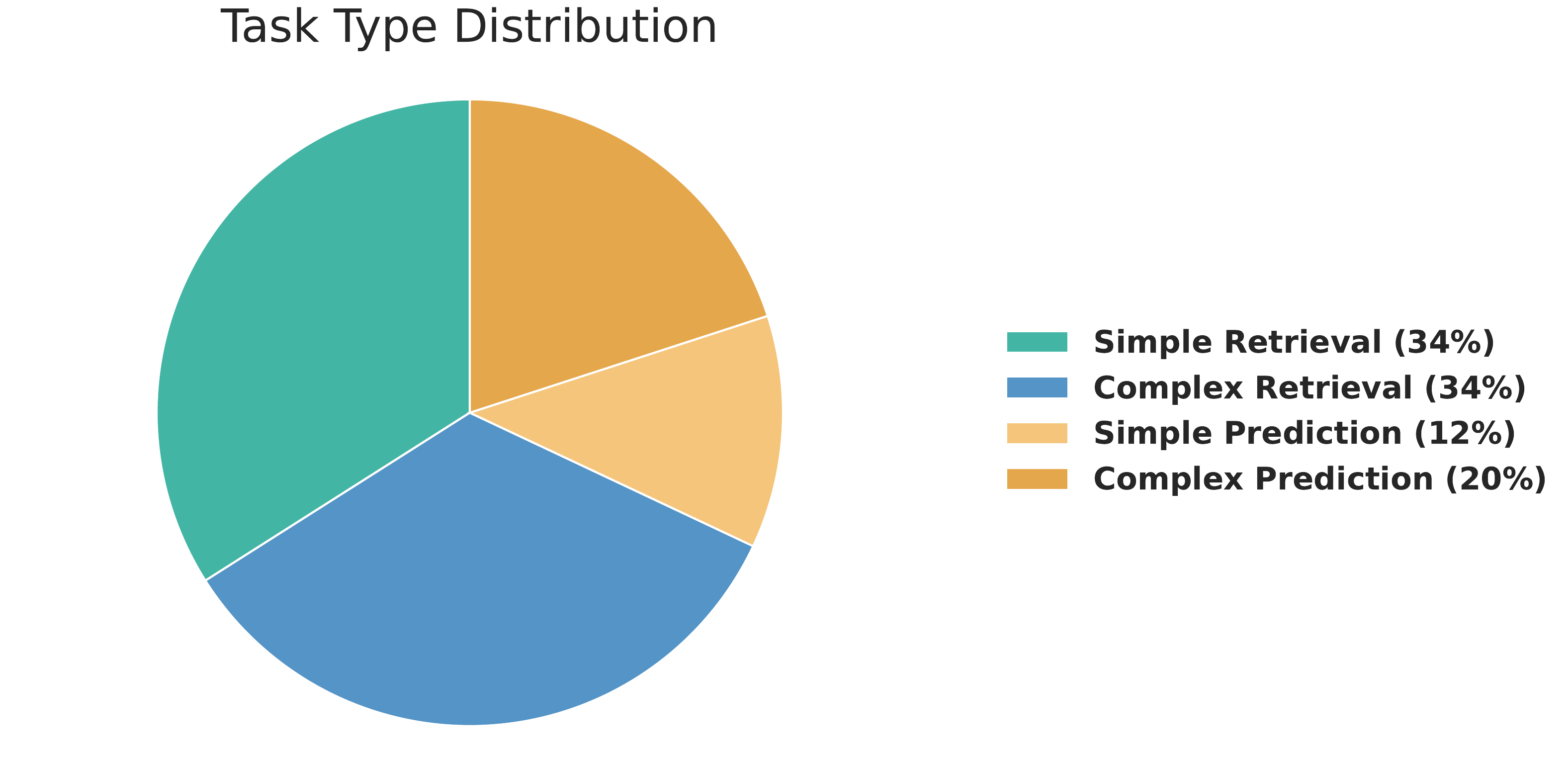}
        \caption{Distribution of Task Types (N=50).}
        \label{fig:task_dist}
    \end{subfigure}
    \hfill
    \begin{subfigure}{0.48\textwidth}
        \includegraphics[width=\linewidth]{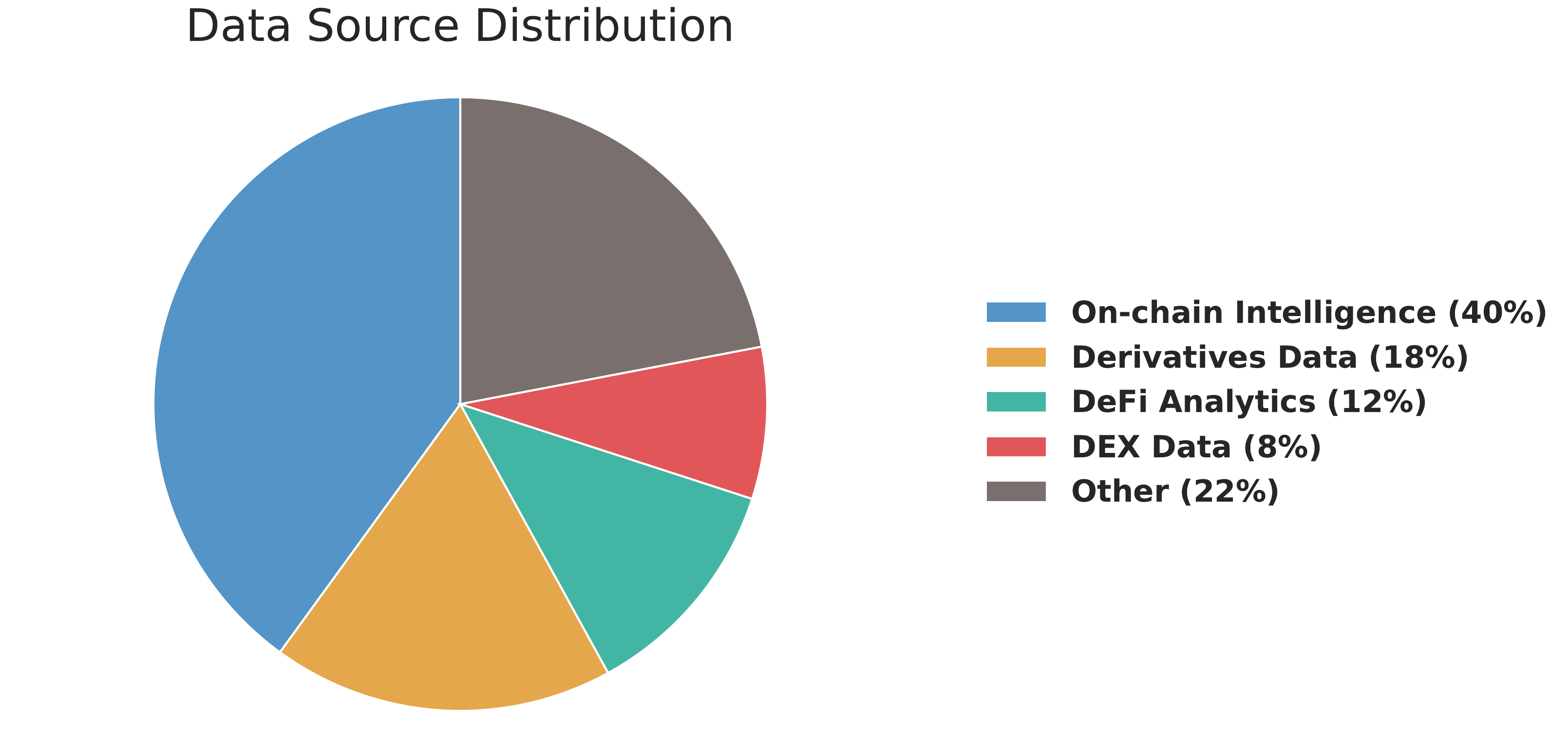}
        \caption{Distribution by Data Source Category.}
        \label{fig:platform_dist}
    \end{subfigure}
    \caption{Statistics of the CryptoBench dataset between October $12^{th}$ to November $11^{th}$.}
    \label{fig:dataset_stats}
\end{figure*}
CryptoBench is engineered from the ground up to serve as a high-fidelity proxy for the complex, dynamic, and adversarial environment of professional cryptocurrency analysis. Its architecture is not a collection of arbitrary tasks, but a carefully structured evaluation system designed to probe the core competencies required for expert-level performance. This design is guided by three foundational principles that ensure its relevance, rigor, and longevity: \textbf{Task Professionalism \& Diversity}, a commitment to being \textbf{Dynamic and Evolving}, and \textbf{Broad Platform Coverage}.

\begin{figure*}[t]
    \centering
    \includegraphics[width=0.9\textwidth]{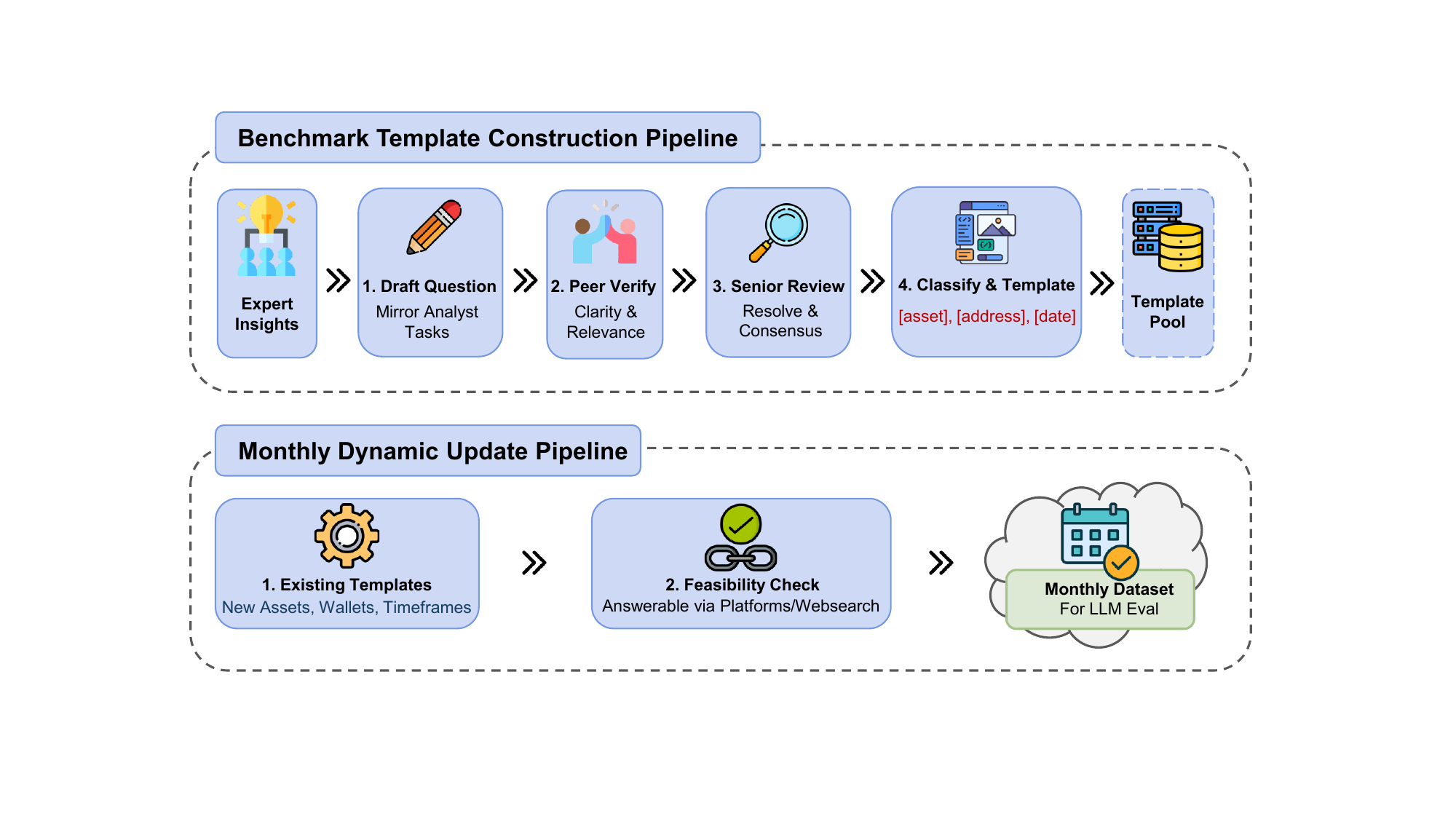}
    \caption{The CryptoBench Dataset Construction and Dynamic Update Pipeline. The top panel illustrates the rigorous multi-stage verification protocol for creating question templates. The bottom panel shows the monthly process for generating fresh, solvable questions from the template pool to ensure the benchmark's timeliness and relevance.}
    \label{fig:construction_pipeline}
\end{figure*}
\subsection{Design Principles}
The design of CryptoBench is guided by three core principles where Table \ref{tab:illustrative_examples} shows the illustrative examples:

\begin{table*}[t]
\centering
\caption{Illustrative Examples of Core Assessment Capabilities in CryptoBench.}
\label{tab:illustrative_examples}
\begin{tabular}{@{}l p{0.72\textwidth}@{}}
\toprule
\textbf{Capability} & \textbf{Abstracted Question Example} \\ \midrule
Real-time Factual Retrieval &
``Does address \texttt{[address\_hash]} currently hold a `Key Opinion Leader' designation according to our on-chain intelligence feed?'' \\ \addlinespace

Historical Data Aggregation &
``What were the opening and closing market capitalizations for token \texttt{[token\_address]} on \texttt{[date]}?'' \\ \addlinespace

Cross-Entity Correlation &
``Compare the monthly returns of \texttt{[Asset A]} and \texttt{[Asset B]} for \texttt{[Quarter, Year]} and identify the outperformer.'' \\ \addlinespace

Behavioral Pattern Inference &
``Analyze the trading history of address \texttt{[address\_hash]} over the past 7 days to forecast its likely number of buy and sell transactions for the upcoming week.'' \\ \addlinespace

Short-term Market Prediction &
``Based on current market volatility, which three crypto assets are most likely to experience the highest liquidation volumes in the next 24 hours?'' \\ 
\bottomrule
\end{tabular}
\end{table*}

\begin{figure*}[t]
    \centering
    \includegraphics[width=0.8\textwidth]{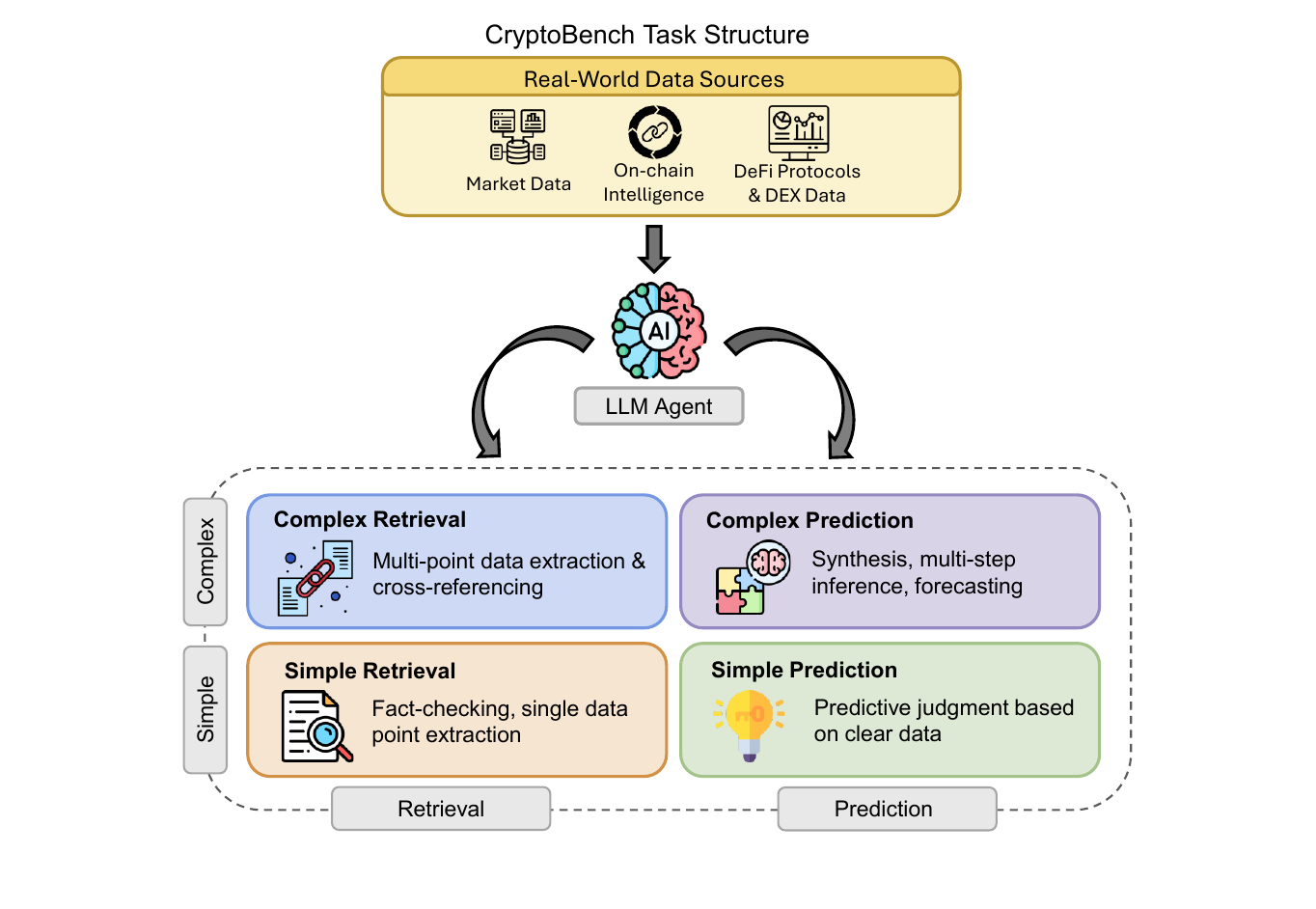}
    \caption{The CryptoBench Four-Quadrant Task Classification System. Tasks are categorized along two axes: Complexity (Simple vs. Complex) and Cognitive Demand (Retrieval vs. Prediction), providing a granular view of agent capabilities.}
    \label{fig:task_structure}
\end{figure*}

\textbf{Task Professionalism \& Diversity:} To ensure ecological validity, every question in CryptoBench is meticulously crafted by a committee of crypto-native professionals, including DeFi analysts, on-chain intelligence investigators, and quantitative traders. This grounds the benchmark in the practical realities of the industry, moving beyond academic exercises to mirror the day-to-day queries and analytical workflows that define these roles. The tasks are deliberately diverse, designed to test a wide spectrum of cognitive and operational skills. They cover critical domains such as:
(i) \textbf{On-chain Intelligence}: Requiring agents to analyze a wide range of on-chain phenomena, such as the activities of whale and Key Opinion Leader (KOL) wallets, user profitability and win rates, patterns of token accumulation or distribution, and security risks like phishing wallets or potential ``Rat Trading''.
(ii) \textbf{Market Data Analysis}: Tasks involving the retrieval of historical data and the interpretation of macro-level real-time feeds, such as aggregate liquidations, open interest, funding rates, and long/short ratios.
(iii) \textbf{DeFi Protocol \& Oracle Analysis}: Questions that demand navigation of specific protocol dashboards to find information like Total Value Locked (TVL), and comparing Total Value Secured (TVS) across different oracles.
(iv) \textbf{DEX \& Derivatives Analytics}: Probing an agent's ability to interpret data from Decentralized Exchanges (DEXs) and derivatives platforms, such as funding rates, open interest, and liquidation levels.
(v) \textbf{MEV and AI-driven Signals}: Advanced tasks requiring the analysis of MEV opportunities on specific chains and the interpretation of AI-generated trading signals from specialized platforms.
These diverse domains require a holistic evaluation, preventing a model from achieving a high score by excelling in only one narrow skill, such as simple data retrieval. Instead, it measures an agent's versatility and its ability to function as a genuine analytical co-pilot, being able to combine general agent abilities with domain-specific knowledge.

\textbf{Dynamic and Evolving:} The cryptocurrency market is defined by its relentless pace of innovation and change; protocols emerge, narratives shift, and token standards evolve in a matter of weeks. A static benchmark in such an environment would quickly become outmoded and susceptible to contamination, where models are simply trained on the answers. To combat this, CryptoBench is designed as a \textbf{``living'' benchmark}. Its questions are built on a templated framework that allows for rapid and systematic updates. Core entities within questions—such as token tickers, wallet addresses, transaction hashes, and timeframes—are treated as variables that can be periodically refreshed. We have established a roadmap for \textbf{monthly updates} to the benchmark. These updates will not only refresh the data points in existing questions but will also introduce entirely new tasks that reflect emerging market narratives and DeFi primitives (e.g., questions related to restaking, modular blockchains, or new oracle designs). This dynamic structure serves two critical purposes: 1) it ensures the benchmark's \textbf{long-term relevance} by keeping pace with the market, and 2) it provides strong \textbf{resistance to contamination}, as memorizing past question-answer pairs will offer diminishing returns. This forces agents to demonstrate genuine, generalizable capabilities rather than pattern-matching against a fixed dataset.

\textbf{Broad Platform Coverage:} No professional crypto analyst operates using a single source of information. Expert analysis is an act of synthesis, requiring the integration of data from a mosaic of specialized, often competing, platforms. CryptoBench explicitly replicates this reality. Tasks are designed to compel agents to navigate, query, and extract data from a wide array of real-world, live platforms essential to the professional workflow. This includes: leading on-chain intelligence aggregators for wallet profiling, transaction tracing, and address-level data; authoritative market data providers for price feeds, real-time trading data, user activity metrics, and historical data; the native dashboards and analytics pages of major DeFi protocols; DEX aggregators and information sites for swap rates and liquidity data; and specialized platforms for futures and derivatives data, such as macro real-time liquidations, funding rates, and open interest. This principle tests more than just an agent's web browsing ability; it evaluates its capacity for \textbf{tool orchestration}. Can the agent identify the most reliable source for a given piece of information? Can it adapt to different user interfaces and data formats? And most importantly, can it correctly synthesize conflicting or complementary information from multiple sources to arrive at a correct conclusion? By forcing interaction with the actual tools of the trade, CryptoBench provides a far more realistic assessment of an agent's practical utility than benchmarks confined to sandboxed or simplified environments.

\subsection{Task Design}
To move beyond a monolithic, single-score evaluation, CryptoBench employs a granular classification system that provides a nuanced, multi-dimensional view of agent capabilities. Recognizing that the tasks of a financial analyst are not uniform, we categorize each question into a \textbf{four-quadrant system} (visualized in Figure~\ref{fig:task_structure}). This system is defined by two axes: the primary cognitive demand (\textbf{Retrieval vs. Prediction}) and the operational complexity (\textbf{Simple vs. Complex}). This framework allows us to precisely diagnose an agent's strengths and weaknesses, distinguishing, for example, between an agent that excels at finding facts but fails at synthesizing them and one that can make logical predictions but struggles with accurate data extraction.

\textbf{Simple Retrieval (SR):} This quadrant assesses the most fundamental capability of an agent: its ability to locate and extract a single, discrete piece of information from a specified or implied source. These tasks are akin to fact-checking or targeted data lookup. Success requires accurate navigation, parsing of a webpage or data feed, and precise extraction of the target data point without further manipulation.
\textit{Example:} ``What is the current 24-hour trading volume for the ETH/USDC pair on Uniswap V3 as reported by its official analytics page?''

\textbf{Complex Retrieval (CR):} This quadrant builds upon simple retrieval by requiring the agent to find and consolidate multiple related data points. The task may involve filtering a table, iterating through a list, or cross-referencing information within a single, complex source.
\textit{Example:} ``From the list of a protocol's top ten holders on Etherscan, identify the two wallets labeled as `KOL' and retrieve their respective token balances and the percentage of total supply they hold.''

\textbf{Simple Prediction (SP):} This quadrant marks the transition from data extraction to data interpretation. Tasks here require the agent to perform a basic inference, calculation, or predictive judgment based on one or a few readily available data points.
\textit{Example:} ``Token A has a major token unlock scheduled for next week, releasing 10\% of its circulating supply to early investors. Based on this single event, is the short-term price pressure more likely to be bullish or bearish?''

\textbf{Complex Prediction (CP):} This is the most demanding quadrant, designed to simulate the comprehensive analytical tasks performed by senior analysts. These questions require multi-step inference, the synthesis of information from multiple, potentially conflicting sources, and the formulation of a forecast or strategic recommendation.
\textit{Example:} ``Analyze the trading history of address \texttt{0x...} over the past 7 days, considering its average trade size, frequency, and choice of assets, to forecast its likely number of buy and sell transactions for the upcoming week.''

This four-quadrant system provides a powerful diagnostic tool. An agent that scores highly on SR and CR but poorly on SP and CP is a competent data retriever but an ineffective analyst. Conversely, an agent that scores well on prediction tasks but fails at retrieval may be prone to ``hallucinating'' plausible but factually incorrect conclusions. A truly expert-level agent must demonstrate proficiency across all four quadrants.

\subsection{Dataset Construction and Quality Control}
The credibility of any benchmark rests on the quality of its dataset. Recognizing this, the construction of CryptoBench was a meticulous, multi-stage process designed to produce questions and answers of the highest clarity, relevance, and verifiability. This process was a collaboration involving a dedicated team of crypto-native professionals, governed by a rigorous peer-verification protocol (visualized in Figure~\ref{fig:construction_pipeline}) designed to eliminate ambiguity and ensure every question represents a valid, real-world analytical challenge.

\textbf{Expert-Led Curation:}
The foundation of CryptoBench is human expertise. The benchmark was constructed by a curated team of crypto-native professionals with deep, practical experience in the field. This team included: \textbf{DeFi Analysts}: Experts accustomed to evaluating protocol health, yield farming strategies, and tokenomic models; \textbf{On-Chain Intelligence Investigators}: Specialists skilled in using block explorers and advanced analytics platforms to trace fund flows, identify sophisticated actors (whales, market makers), and detect anomalous behavior; and \textbf{Quantitative Traders}: Professionals who rely on precise, timely data to build and execute trading models. This diversity of expertise ensures that the questions are not merely academic but are grounded in the tangible, high-stakes tasks that define professional work in the cryptocurrency space.

\textbf{Rigorous Multi-Stage Verification Protocol:}
To ensure each question is unambiguous, relevant, and has a clear, verifiable answer, we implemented a stringent three-stage validation protocol. A question was only included in the final dataset if it achieved unanimous consensus across all stages:
(1) \textbf{Stage 1: Generation and Ground-Truthing.} An initial expert (the ``Author'') drafts a question based on a real-world scenario. Crucially, the Author must also provide a detailed ``gold-standard'' answer, a step-by-step solution path describing how the answer can be obtained, and direct links to the primary sources required. This initial ground-truthing ensures that every question is solvable from the outset.
(2) \textbf{Stage 2: Adversarial Validation.} The drafted question is then passed to a second expert (the ``Validator'') who has no knowledge of the pre-compiled answer. The Validator's role is to act as an independent, adversarial agent. They attempt to answer the question from scratch, documenting their own process. They then compare their result to the Author's. This stage is designed to identify potential failure points including clarity, source reliability, answer stability, and feasibility.
(3) \textbf{Stage 3: Final Adjudication and Standardization.} If any discrepancy arises between the Author and the Validator, the question is escalated to a third, senior expert (the ``Adjudicator''). The Adjudicator reviews the arguments from both sides, makes a final determination on the question's validity, and may propose modifications to improve its clarity. This stage also serves to standardize phrasing, difficulty ratings, and formatting across the entire benchmark, ensuring consistency.

\textbf{Systematic Mitigation of Ambiguity:}
Financial and on-chain data are notoriously noisy and can be presented differently across platforms. A primary focus of our quality control was the systematic elimination of ambiguity. We implemented several specific measures: \textbf{Source Specificity}: Questions explicitly name the required data source whenever possible (e.g., ``as reported on the protocol's official dashboard'', ``according to CoinGecko's API'', ``on the Etherscan token holder chart''). \textbf{Metric Definition}: We avoid ambiguity by defining metrics precisely. Questions avoid vague terms like ``market size'', ``users'', or ``trading activity''. \textbf{Temporal Bounding}: For questions involving time-sensitive data, clear timeframes are provided (e.g., ``in the last 24 hours'', ``during Q1 2024''). For highly volatile, real-time numerical answers, we establish clear tolerance ranges for evaluation (e.g., a $\pm 5\%$ margin for TVL figures) to account for minor fluctuations during the evaluation period.

\begin{figure*}[!t]
    \centering
    \begin{subfigure}{0.48\textwidth}
        \includegraphics[width=\linewidth]{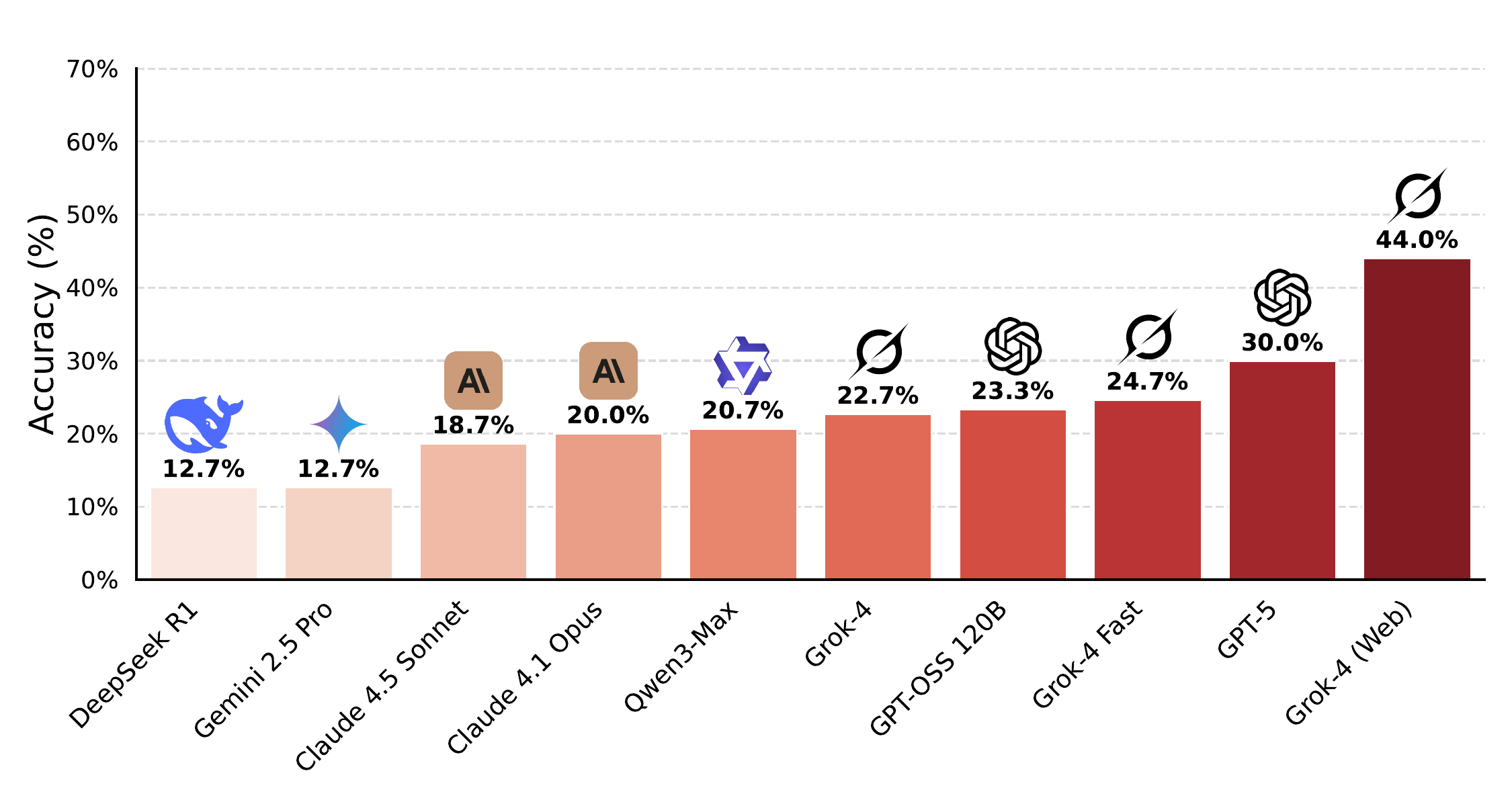}
        \caption{LLM Evaluation Scores.}
        \label{fig:llm_eval_scores}
    \end{subfigure}
    \hfill
    \begin{subfigure}{0.48\textwidth}
        \includegraphics[width=\linewidth]{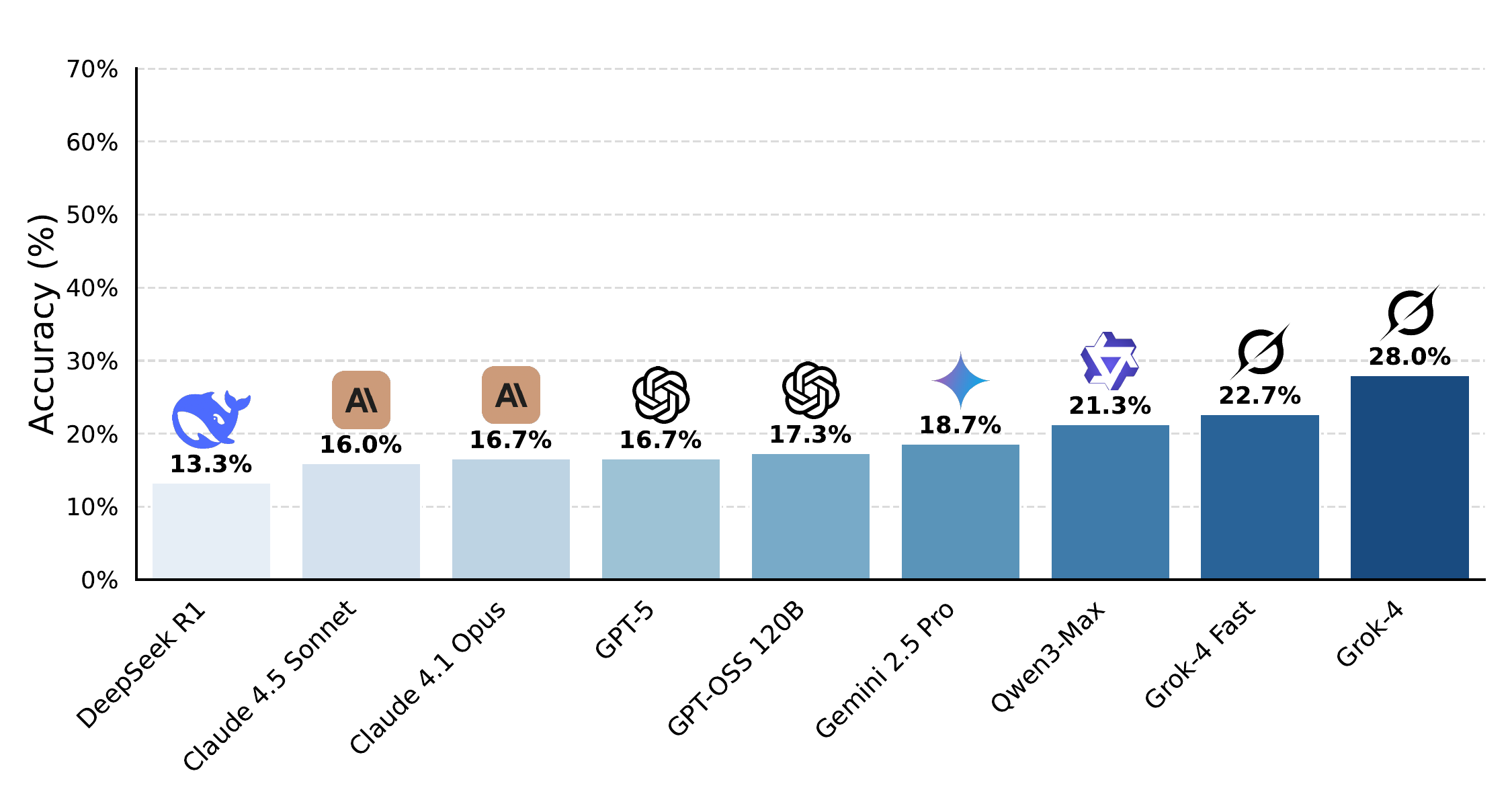}
        \caption{SmolAgent Evaluation Scores.}
        \label{fig:smolagent_eval_scores}
    \end{subfigure}
    \caption{Overall performance comparison between October $12^{th}$ and November $11^{th}$.}
    \label{fig:overall_performance}
\end{figure*}
\section{Experiments}
\label{sec:experiments}
To rigorously assess the capabilities of modern LLM agents against the challenges posed by CryptoBench, we designed a comprehensive experimental setup. This section details the selection of models, the sophisticated protocol used for evaluation, and the metrics employed to derive meaningful insights from the results.

\subsection{Evaluated Models and Agentic Framework}
To establish a robust and comprehensive baseline, we evaluated ten state-of-the-art LLM agents. Our selection was curated to represent the cutting edge of both commercially available, closed-source models and leading open-source alternatives, all of which are recognized for their advanced prediction and tool-use capabilities. The models are: \textbf{Grok-4 (Web)}, \textbf{GPT-5} \cite{openai2025gpt5}, \textbf{Grok-4} \cite{xai2025grok4}, \textbf{Grok-4 Fast} \cite{xai2025grok4fast}, \textbf{Qwen3-Max} \cite{yang2025qwen3}, \textbf{Claude 4.1 Opus}, \textbf{DeepSeek R1} \cite{deepseek2025r1}, \textbf{Claude 4.5 Sonnet}, \textbf{Gemini 2.5 Pro} \cite{google2025gemini25}, and \textbf{GPT-OSS 120B}. Except for Grok-4 (Web), which was evaluated via its native web interface, all other models were accessed through their online versions on OpenRouter.

This selection provides a diverse cross-section of the current AI landscape, from flagship models designed for complex, multi-turn predictive analysis to more agile, performance-optimized variants. To ensure a fair and standardized comparison, each model was integrated into an identical agentic framework. This framework equipped each LLM with a single, powerful tool: a \textit{web-browsing agent} capable of performing fundamental actions such as navigating to a URL, searching the web with a query, and extracting textual content from a webpage \cite{comanici2025gemini}. No model was given preferential access to proprietary APIs or specialized tools; their performance depended solely on their ability to formulate a plan for the task and effectively wield this general-purpose browsing tool to interact with the live, real-world platforms required by CryptoBench. This setup mirrors a realistic scenario where an agent must operate on the open internet to solve domain-specific problems.

\subsection{Evaluation Protocol}
Evaluating complex, open-ended responses in a dynamic domain like finance requires a more nuanced approach than simple string matching or binary correct/incorrect scoring. Therefore, we employ a sophisticated evaluation protocol centered on an \textit{LLM-as-a-Judge} framework \cite{zheng2023judging}, which combines the scalability of automated evaluation with the nuanced judgment of human expertise.

Our protocol is guided by a detailed, multi-level rubric that scores each agent's final response on a scale of 0 to 3. This granular scoring system is designed to capture degrees of correctness and provide insight into failure modes:
\textbf{Score 3} (Completely Correct): The agent provides a factually accurate answer that directly and fully addresses the question. All numerical values are within the acceptable tolerance, and the conclusion is sound.
\textbf{Score 2} (Mostly Correct): The agent's response is substantially correct and demonstrates a correct approach, but contains minor inaccuracies. This could be a small miscalculation, a partially incomplete list, or a correct conclusion derived from slightly flawed premises.
\textbf{Score 1} (Partially Correct): The agent shows some understanding of the task and may have successfully retrieved some relevant information, but the final answer is incorrect. This score acknowledges a valid attempt but an ultimate failure in execution or prediction.
\textbf{Score 0} (Incorrect): The agent's response is completely wrong, irrelevant to the question, a clear hallucination, or the agent fails to produce any answer at all.

To account for the high volatility of real-time financial data, we apply a \textbf{$\pm$5\% tolerance margin} for any question requiring a numerical answer that is subject to market fluctuations (e.g., prices, TVL, market cap). This prevents an agent from being unfairly penalized for minor discrepancies caused by data-fetching latency. The LLM-judge is provided with the original question, the agent's full response trace, the ground-truth answer, and this detailed rubric to ensure consistent and fair scoring across all models.

\subsection{Evaluation Metrics}
To aggregate the fine-grained scores from our rubric into a clear and interpretable measure of performance, we compute the \textbf{Average Success Rate}. This metric provides a normalized score for each model, reflecting its overall proficiency on the benchmark. It is calculated by summing the scores awarded to a model across all questions and dividing by the maximum possible score, then expressing the result as a percentage.

The formula is as follows:
\[
\text{Average Success Rate} = \left( \frac{\sum_{i=1}^{N} \text{score}_i}{N \times 3} \right) \times 100\%.
\]
where $N$ is the total number of questions, and 3 is the maximum possible score for a single question.

\begin{figure*}[t]
    \centering
    \begin{subfigure}{0.48\textwidth}
        \includegraphics[width=\linewidth]{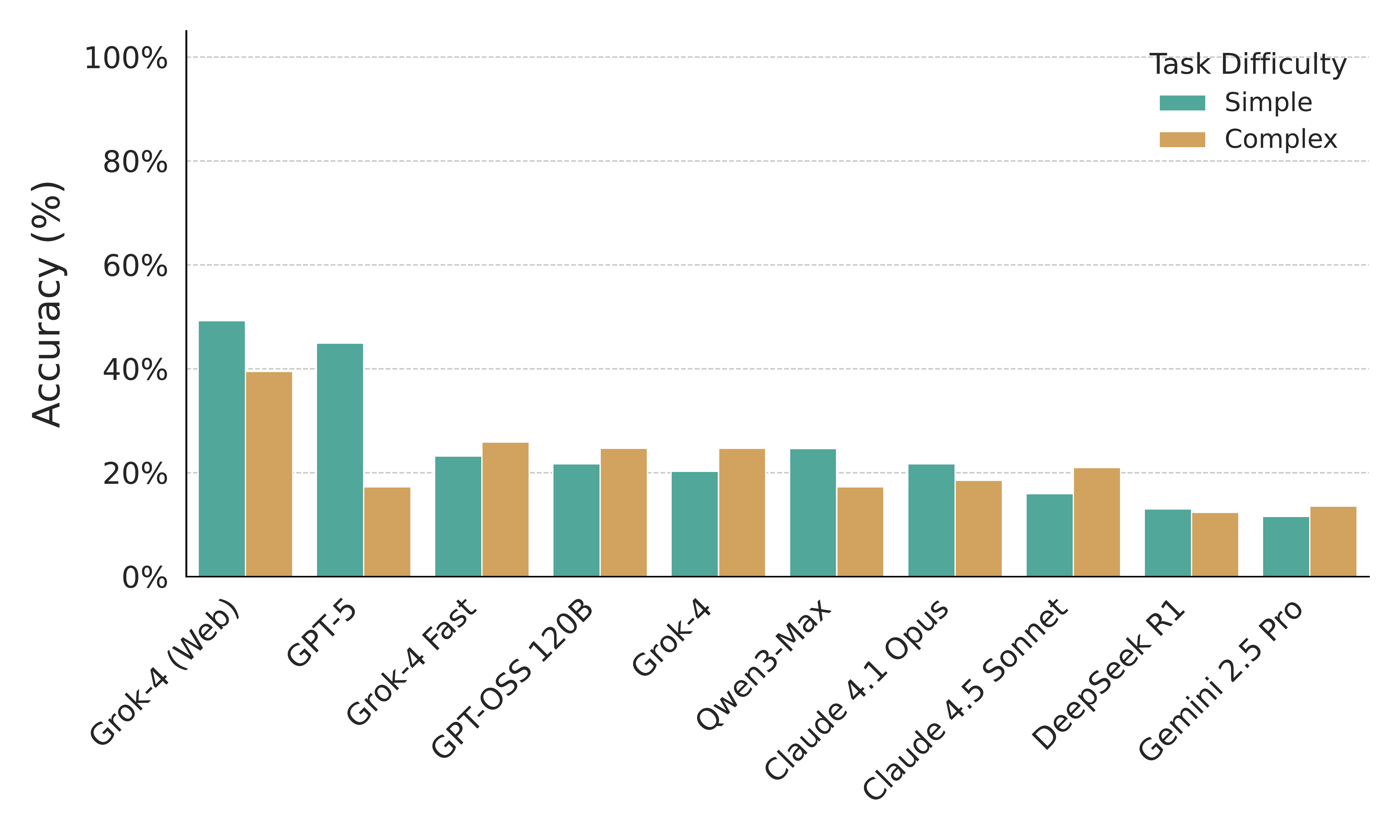}
        \caption{Performance by Task Difficulty.}
        \label{fig:perf_by_difficulty}
    \end{subfigure}
    \hfill
    \begin{subfigure}{0.48\textwidth}
        \includegraphics[width=\linewidth]{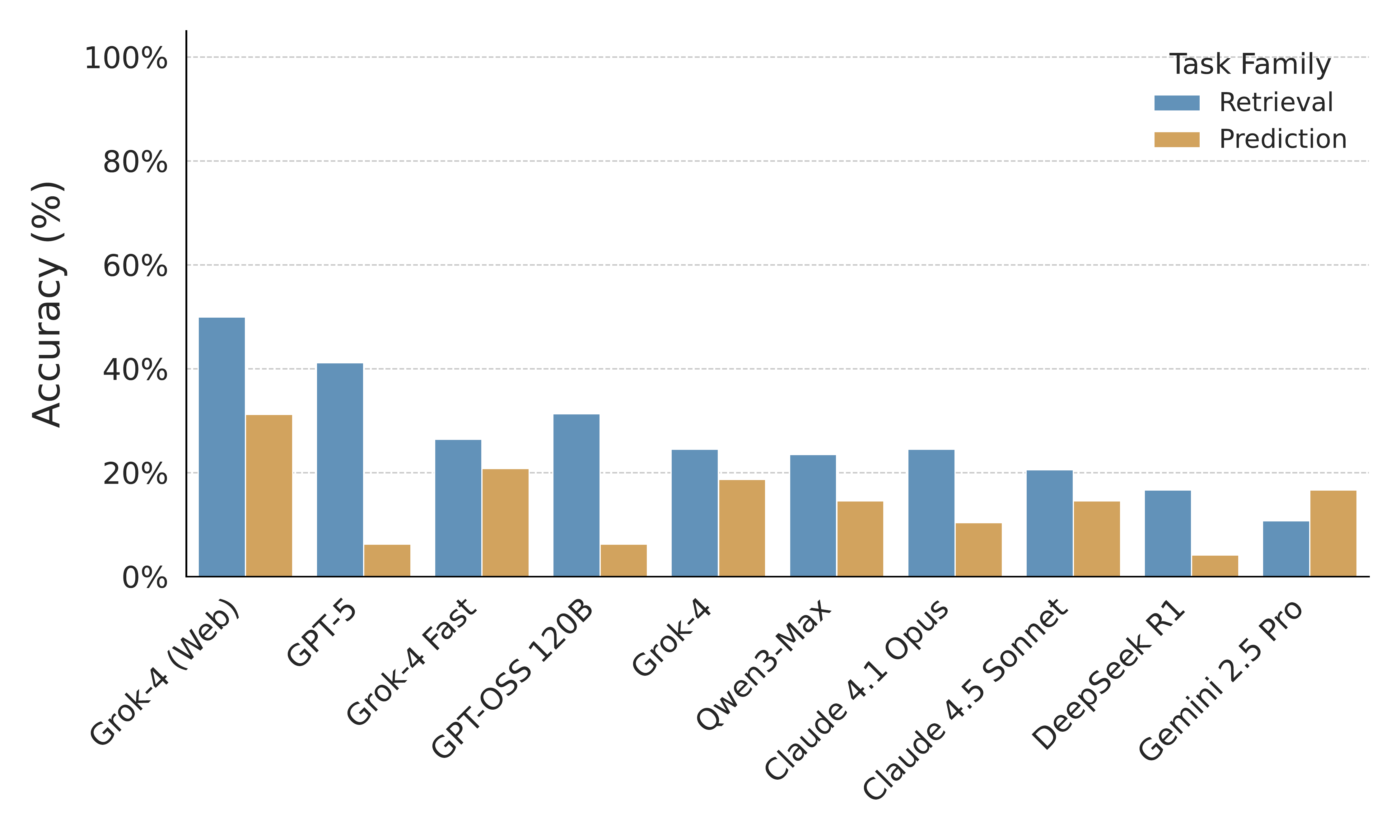}
        \caption{Performance by Task Family.}
        \label{fig:perf_by_family}
    \end{subfigure}
    \caption{Performance breakdown between October $12^{th}$ and November $11^{th}$. (a) Comparison of model accuracy on Simple versus Complex tasks. (b) Comparison of accuracy on Retrieval versus Prediction tasks, revealing distinct model strengths.}
    \label{fig:performance_breakdown}
\end{figure*}

Our experiments provide a comprehensive assessment of current LLM capabilities on expert-level cryptocurrency tasks. Overall, the results reveal a clear performance hierarchy across models, a substantial gap between retrieval and prediction abilities, and a strong dependence of outcomes on the evaluation setting (direct prompting versus agentic execution).

\subsection{Overall Performance: LLM vs. Agentic Framework}

Across all evaluated models, direct LLM evaluation and agent-based evaluation yield markedly different performance profiles. In direct evaluation, a small number of models substantially outperform the rest, indicating large variance in raw task-solving ability. When deployed within an agentic framework, overall accuracy generally decreases, but relative model rankings shift, suggesting that agentic execution introduces additional challenges related to planning, tool use, and error propagation.

Notably, while some models experience performance degradation under the agentic setting, others exhibit comparatively stronger robustness, indicating that raw model capability does not directly translate into effective agentic performance. Detailed quantitative results are provided in Appendix~\ref{sec:appendix_overall_performance}.

\subsection{Effects of Task Difficulty and Task Family}

Performance consistently declines as task complexity increases, but the magnitude of this degradation varies significantly across models. While top-performing models retain reasonable accuracy on complex tasks, others experience sharp drops, highlighting limitations in multi-step reasoning and synthesis.

A more pronounced divide emerges when comparing retrieval and prediction tasks. Across nearly all models, retrieval tasks are handled substantially better than prediction tasks, revealing a systematic retrieval--prediction imbalance. In several cases, models that perform strongly on retrieval exhibit near-random performance on predictive tasks, underscoring fundamental weaknesses in inferential reasoning under uncertainty. Figure~\ref{fig:performance_breakdown} illustrates model performance decomposed along these two complementary axes.

\subsection{Performance Across Crypto-Specific Domains}

Analysis across macro-level cryptocurrency domains shows that models perform best on tasks requiring relatively general knowledge or less specialized data access. In contrast, tasks involving deep interaction with niche, professional-grade data sources—such as decentralized exchange data and on-chain intelligence—remain challenging for most models. This suggests that current agents struggle to reliably interface with specialized financial analytics platforms. Domain-wise performance visualizations are included in Appendix~\ref{sec:appendix_macro_categories}.

\subsection{Granular Analysis by Task Quadrant}

A four-quadrant analysis further highlights stark asymmetries in model capabilities. Most models excel in simple retrieval but exhibit dramatic performance collapse in both simple and complex prediction. Even high-performing models display limited competence in predictive reasoning, indicating that failures are not merely due to task difficulty but reflect deeper architectural limitations. Detailed quadrant-level results are provided in Appendix~\ref{sec:appendix_task_quadrants}.

\subsection{Qualitative Failure Mode Analysis}

Manual inspection of incorrect responses reveals several recurring failure patterns. Agents frequently rely on non-authoritative or outdated sources, retrieve stale cached information in rapidly changing markets, fail to correctly integrate multiple retrieved data points, or hallucinate unsupported predictions. These failure modes are especially prevalent in predictive tasks and align with prior observations on the limitations of LLM forecasting under real-world uncertainty. Representative qualitative examples and detailed analyses are provided in Appendix~\ref{sec:appendix_failure_modes}.
\vspace{-0.2cm}

\section{Conclusion}
\label{sec:conclusion}
\vspace{-0.2cm}

In this paper, we introduced CryptoBench, a novel and challenging benchmark designed to evaluate LLM agents on real-world cryptocurrency analysis tasks. Our comprehensive evaluation reveals that the current generation of agents, despite their impressive general capabilities, fall significantly short of expert-level performance in this demanding domain. We identified a surprising inversion in performance on retrieval versus prediction tasks, a near-total failure at a specific difficulty threshold, and a consistent inability to handle complex on-chain analysis. CryptoBench demonstrates that the next frontier for financial LLM agents is not just better tool use, but the development of domain-specific prediction architectures capable of navigating adversarial and rapidly evolving information landscapes. We release CryptoBench as a vital, evolving resource to catalyze the development of more robust, reliable, and expert-level financial agents. A comprehensive discussion is provided in  Appendix \ref{sec:discussion}.

\section*{Impact Statement}

This paper presents work whose goal is to advance the field of Machine Learning by providing a benchmark for evaluating LLM agents in the cryptocurrency domain. While CryptoBench enables rigorous evaluation of AI capabilities in financial analysis, we note that improvements in such agents could have dual-use implications, potentially aiding both legitimate analysts and bad actors. We encourage responsible development and deployment of AI systems in financial contexts.

\bibliography{references}
\bibliographystyle{icml2026}

\newpage
\appendix
\onecolumn
\section*{Appendix}
\appendix
\section{Additional Background and Analysis}
\label{sec:appendix}

\subsection{Characteristics of the Cryptocurrency Domain}
\label{sec:crypto_characteristics}

Cryptocurrency markets represent an extreme instantiation of financial complexity. Unlike traditional financial markets, crypto markets operate continuously without centralized oversight and exhibit exceptionally high data velocity, with blockchain transactions confirming every few seconds and market data updating in real time.

The domain is further characterized by high levels of noise and adversarial behavior. Valuable signals are often embedded within conflicting narratives across social media platforms such as X and Telegram, anonymous developer communications, governance forums, and security audit reports. In addition, decentralized smart contracts and automated market makers introduce novel attack surfaces, including sophisticated market manipulation and coordinated misinformation campaigns.

Finally, cryptocurrency analysis relies heavily on unstructured and semi-structured data sources. Analysts must simultaneously interpret on-chain transaction graphs, technical whitepapers, protocol documentation, and open-source code repositories. These properties jointly make cryptocurrency a uniquely challenging environment for evaluating agentic reasoning and decision-making.

\subsection{Motivating Example: Professional Crypto Analyst Tasks}
\label{sec:motivating_example}

Unlike purely informational tasks, real-world cryptocurrency analysis is inherently analytical and forward-looking. A typical analyst query may involve assessing market participant behavior and predicting future price dynamics, for example:

\begin{quote}
``Analyze the on-chain activity of the top five holders of Token X over the past 48 hours to determine whether whale accumulation or distribution is occurring.''
\end{quote}

Answering such a query requires integrating multiple data sources, including on-chain transaction histories, wallet clustering heuristics, exchange inflow and outflow analysis, and temporal behavior patterns. This type of reasoning extends beyond document retrieval or factual extraction and demands predictive judgment under uncertainty.

\subsection{Limitations of Existing Agent Benchmarks}
\label{sec:benchmark_limitations}

General-purpose agent benchmarks provide valuable insights into broad agent capabilities but fall short in the cryptocurrency domain.

WebArena \cite{zhou2024webarena} evaluates an agent's ability to execute instructions within static web environments but does not assess interaction with real-time, continuously updating financial dashboards or decentralized finance (DeFi) protocols.

GAIA \cite{mialon2023gaia} focuses on factual extraction and verification from news articles but does not evaluate cross-verification against on-chain transaction data or adversarial misinformation, both of which are critical survival skills in crypto markets.

Similarly, AgentBench \cite{liu2023agentbench} emphasizes general tool use and task completion but lacks domain-specific data sources and does not test predictive reasoning under financial uncertainty.

As a result, these benchmarks cannot measure an agent's practical ability to execute real-world cryptocurrency analyst workflows.

\section{Extended Related Work Discussion}
\label{sec:extended_related_work}

\subsection{Comparison with General Agentic Benchmarks}
\label{sec:agentic_benchmark_comparison}

General-purpose agentic benchmarks such as WebArena \cite{zhou2024webarena} and BrowseComp \cite{wei2025browsecomp} primarily evaluate agents' proficiency in web navigation and information retrieval. WebArena, in particular, focuses on generalized browsing tasks but does not require agents to synthesize numerical data from multiple structured financial APIs, which is a core challenge in CryptoBench.

GAIA \cite{mialon2023gaia} targets general AI assistants with diverse real-world tasks, while AgentBench \cite{liu2023agentbench} evaluates agents across multiple environments with an emphasis on tool use and task completion. SWE-bench \cite{jimenez2023swe} tests agents on real-world GitHub issue resolution, emphasizing software engineering skills.

While these benchmarks are valuable for assessing broad agentic capabilities, they are intentionally domain-agnostic and do not capture the specialized data interpretation skills, domain knowledge, and adversarial conditions inherent to financial and cryptocurrency analysis \cite{bigeard2025finance, li2025investorbench}. In contrast, CryptoBench requires agents to interact with professional on-chain data analysis platforms and integrate highly structured and unstructured data sources within a real-world financial setting.

\subsection{Time-Sensitive and Dynamic Benchmarks}
\label{sec:time_sensitive_benchmarks}

The importance of evaluating models under time-sensitive and evolving conditions has been increasingly recognized. FreshQA \cite{vu2024freshllms} introduces a dynamic question-answering benchmark designed to test up-to-date world knowledge. LiveBench \cite{white2024livebench} emphasizes contamination-limited evaluation through automatic updates, while TimE \cite{wei2025time} provides a multi-level benchmark for temporal prediction in real-world scenarios.

CryptoBench extends these approaches by not only requiring access to real-time information but also by functioning as a ``living'' benchmark. Its templated question design enables periodic updates, ensuring that tasks remain relevant and resist being solved through static knowledge memorization. This property is particularly critical in the rapidly evolving cryptocurrency domain.

\subsection{Future Prediction and Forecasting Benchmarks}
\label{sec:future_prediction_benchmarks}

Evaluating the forecasting ability of LLMs has become an active area of research. Prior studies assess model performance on real-world forecasting tasks by comparing predictions against expert human forecasters \cite{lu2025evaluating}. Other benchmarks, such as FutureX \cite{zeng2025futurex} and ForecastBench \cite{karger2024forecastbench}, evaluate forecasting capabilities dynamically over time.

Additional efforts focus on prediction markets and future event prediction. OpenEP \cite{guan2024openep} and NaviTomorrow \cite{nako2025navigating} benchmark future event prediction across broader domains.

CryptoBench differentiates itself by grounding predictive tasks in the cryptocurrency ecosystem, incorporating domain-specific metrics, adversarial dynamics, and market volatility that are not captured by broader forecasting benchmarks \cite{lu2025bizfinbenchbusinessdrivenrealworldfinancial, mateega2025financeqa}.






\section{Extended Experimental Results}
\label{sec:appendix_results}

\subsection{Overall Performance Results}
\label{sec:appendix_overall_performance}
Tables~\ref{tab:llm_eval_detailed} and~\ref{tab:smolagent_eval_detailed} provide the full quantitative breakdown of model performance under direct LLM evaluation and within the SmolAgent framework, respectively. These results support the analysis presented in Section~\ref{sec:experiments}.

\begin{table}[h]
\centering
\caption{Detailed LLM Evaluation Results (Direct Prompting).}
\label{tab:llm_eval_detailed}
\begin{tabular}{@{}lcccc@{}}
\toprule
\textbf{Model} & \textbf{SR} & \textbf{CR} & \textbf{SP} & \textbf{CP} \\ \midrule
Grok-4 (Web) & 54.9 & 45.1 & 35.2 & 28.6 \\
GPT-5 & 58.8 & 32.4 & 6.3 & 5.9 \\
Grok-4 & 38.2 & 28.6 & 18.7 & 15.4 \\
Grok-4 Fast & 42.1 & 26.3 & 14.2 & 11.8 \\
Qwen3-Max & 35.6 & 22.1 & 12.5 & 9.7 \\
Claude 4.1 Opus & 33.8 & 19.4 & 10.8 & 8.2 \\
DeepSeek R1 & 28.4 & 15.2 & 8.6 & 6.1 \\
Claude 4.5 Sonnet & 31.2 & 17.8 & 9.4 & 7.3 \\
Gemini 2.5 Pro & 22.6 & 12.8 & 16.7 & 10.2 \\
GPT-OSS 120B & 40.5 & 24.7 & 11.3 & 8.9 \\
\bottomrule
\end{tabular}
\end{table}

\begin{table}[h]
\centering
\caption{Detailed SmolAgent Framework Evaluation Results.}
\label{tab:smolagent_eval_detailed}
\begin{tabular}{@{}lcccc@{}}
\toprule
\textbf{Model} & \textbf{SR} & \textbf{CR} & \textbf{SP} & \textbf{CP} \\ \midrule
Grok-4 & 42.3 & 32.1 & 22.4 & 18.6 \\
Grok-4 Fast & 38.7 & 28.4 & 18.2 & 14.3 \\
Qwen3-Max & 36.2 & 25.8 & 16.5 & 12.8 \\
Gemini 2.5 Pro & 32.4 & 22.1 & 18.7 & 11.4 \\
GPT-5 & 35.6 & 21.3 & 8.4 & 6.2 \\
Claude 4.1 Opus & 30.8 & 18.6 & 12.3 & 9.1 \\
DeepSeek R1 & 26.4 & 14.8 & 10.2 & 7.5 \\
Claude 4.5 Sonnet & 28.2 & 16.4 & 11.6 & 8.4 \\
GPT-OSS 120B & 34.1 & 20.5 & 9.8 & 7.1 \\
\bottomrule
\end{tabular}
\end{table}

\subsection{Performance by Task Difficulty and Task Family}
\label{sec:appendix_task_breakdown}
This breakdown provides insight into whether performance differences arise primarily from increased reasoning complexity or from fundamental differences in task type. Additional quantitative results and detailed tables supporting the analysis in Section~\ref{sec:experiments} are available upon request.

\subsection{Performance Across Macro Categories}
\label{sec:appendix_macro_categories}
\begin{figure*}[t]
  \centering
  \includegraphics[width=0.8\textwidth]{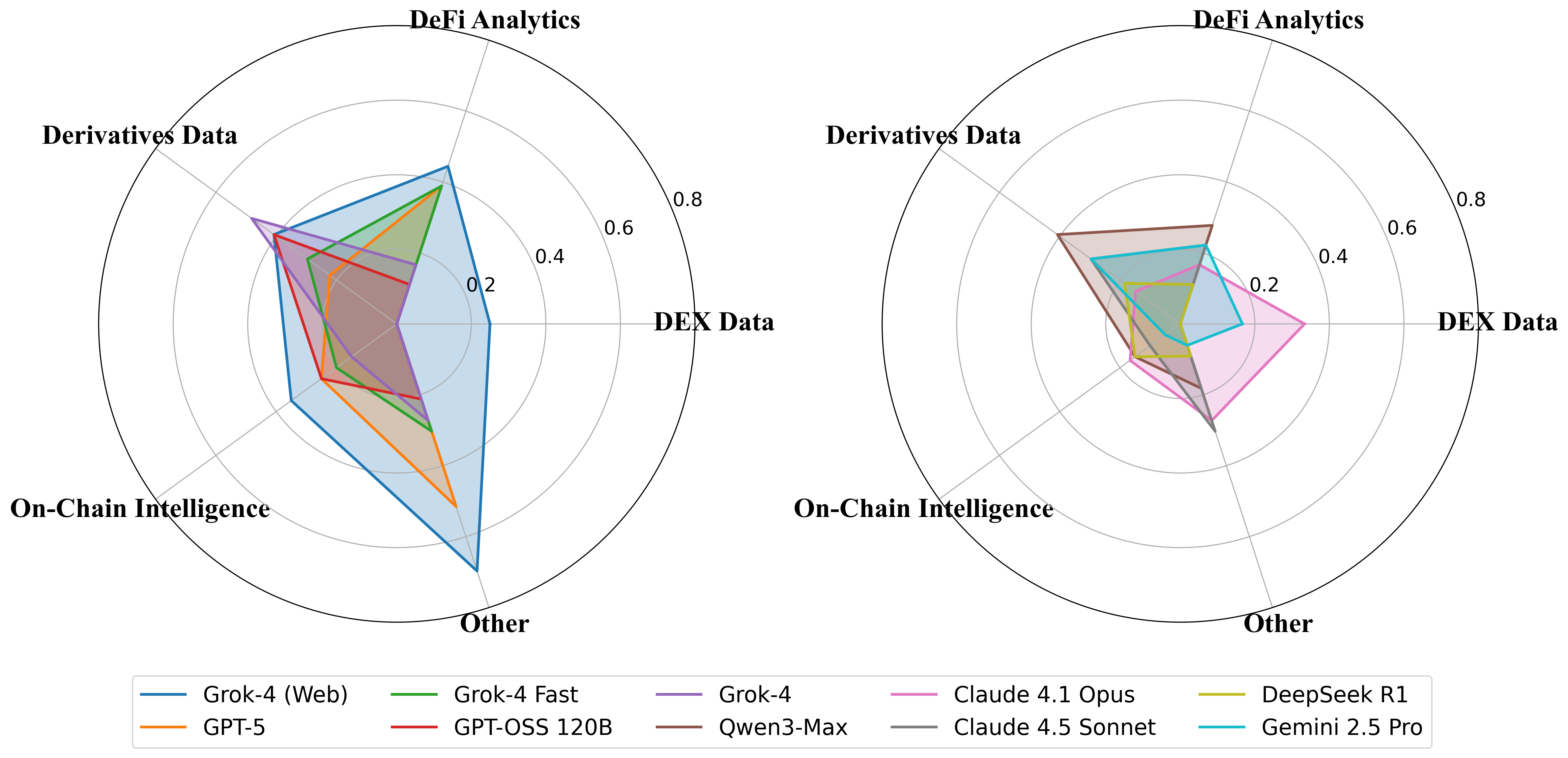}
  \caption{Performance profile by macro category.}
  \label{fig:performance_radar}
\end{figure*}

Figure~\ref{fig:performance_radar} presents a radar-based visualization of model performance across five macro-level cryptocurrency domains: DEX Data, DeFi Analytics, Derivatives Data, On-Chain Intelligence, and Other. Each axis corresponds to a category, and the enclosed area reflects the relative proficiency of a model across these domains.

\subsection{Granular Performance by Task Quadrant}
\label{sec:appendix_task_quadrants}
\begin{figure*}[t]
    \centering
    \begin{subfigure}{0.48\textwidth}
        \includegraphics[width=\linewidth]{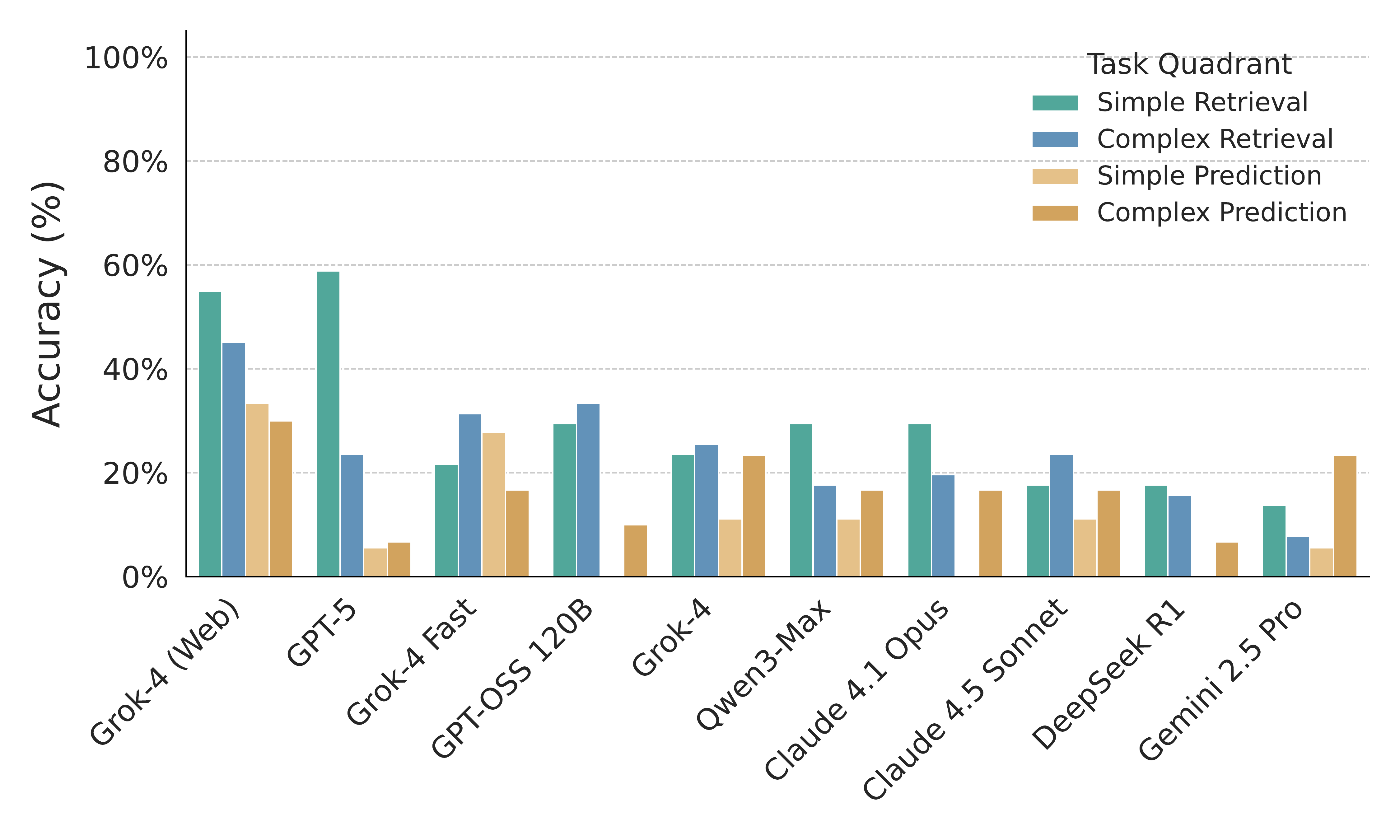}
        \caption{Performance by Task Quadrant.}
        \label{fig:performance_quadrant}
    \end{subfigure}
    \hfill
    \begin{subfigure}{0.48\textwidth}
        \includegraphics[width=\linewidth]{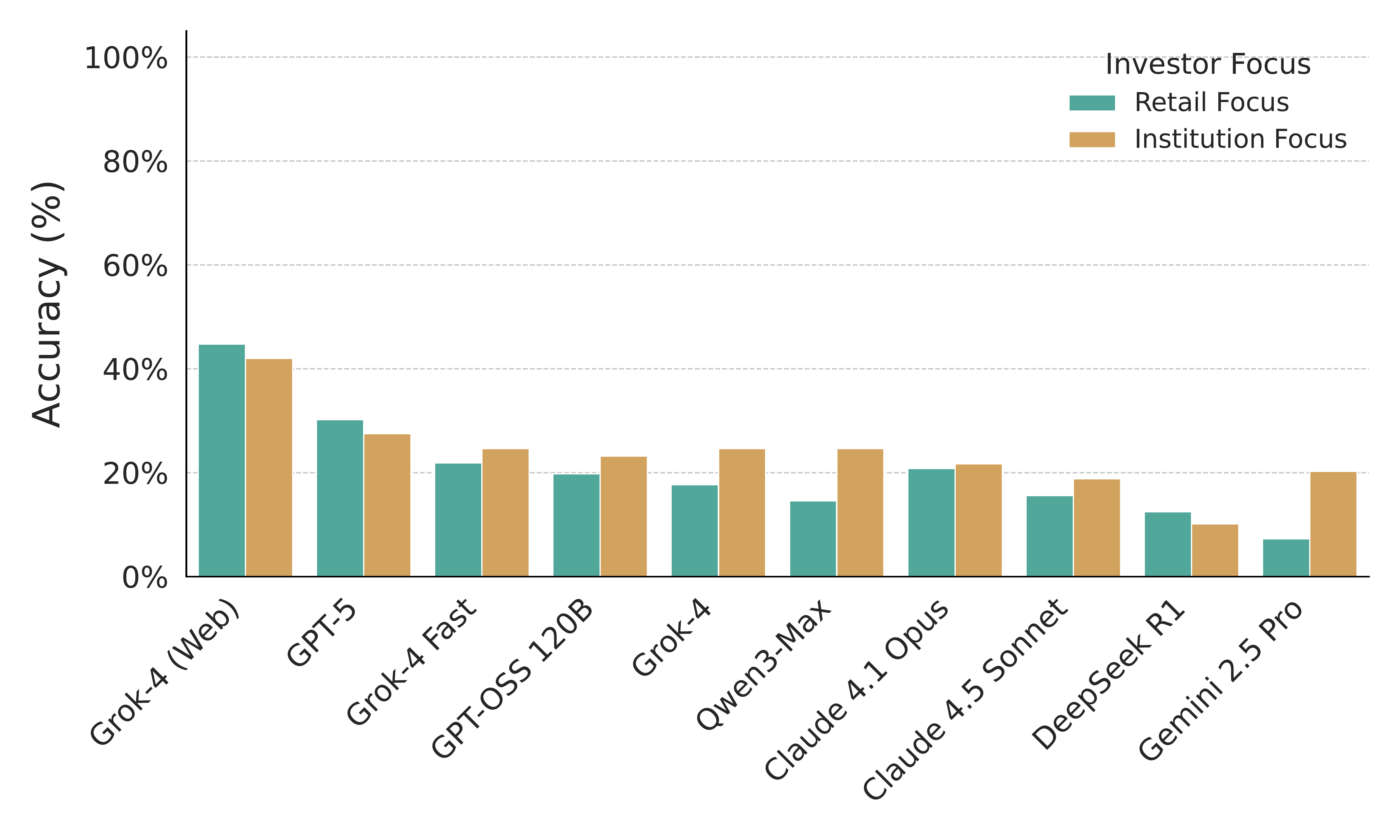}
        \caption{Performance by Investor Focus.}
        \label{fig:perf_investor}
    \end{subfigure}
    \caption{Detailed performance analysis by task quadrant and investor focus.}
    \label{fig:detailed_analysis}
\end{figure*}

Figure~\ref{fig:detailed_analysis} provides a fine-grained breakdown of model performance by task quadrant and investor focus. The quadrant-based analysis decomposes tasks along two axes: retrieval versus prediction, and simple versus complex reasoning, enabling precise identification of capability asymmetries.

\subsection{Qualitative Failure Mode Examples}
\label{sec:appendix_failure_modes}

We identify several recurring failure modes, including shallow source selection, reliance on stale information, integration errors across multiple data points, and hallucinated predictions unsupported by authoritative sources. These qualitative examples further illustrate the systematic weaknesses observed in predictive and real-time reasoning tasks.

\section{Discussion}
\label{sec:discussion}
Our evaluation on CryptoBench reveals a critical gap between the perceived capabilities of modern LLM agents and the practical demands of expert financial analysis. The results underscore three key themes: the fallacy of equating data retrieval with analytical competence, the nuanced relationship between a model's raw intelligence and its agentic effectiveness, and the persistent challenge of interacting with specialized data platforms.

The most significant finding is the stark retrieval-prediction imbalance. Many top models excel at finding discrete facts, creating a ``veneer of competence''. However, this proficiency vanishes when tasks require even simple predictive reasoning. This highlights that current models are better equipped as sophisticated search engines than as analysts, a crucial failure in a domain where value is derived from interpretation, not just information. This suggests that existing training paradigms, focused on factual recall, are insufficient for developing the inferential skills required for financial analysis.

Furthermore, our results show that raw model intelligence does not guarantee effective agentic performance. The shift in rankings between direct LLM evaluation and the agentic framework indicates that the ability to plan and use tools is a distinct skill. A powerful LLM can be bottlenecked by a simple agent framework, suggesting that progress requires co-designing both better models and more sophisticated agentic structures.

Finally, the qualitative analysis reveals a recurring ``last mile'' problem. Agents struggle with source fidelity, often choosing outdated, easily scraped articles over authoritative, real-time dashboards. Their inability to parse the complex, dynamic interfaces of specialized platforms for on-chain and DEX data remains a major barrier. This points to a need for domain-specific tools that go beyond generic web browsing.

\end{document}